University of Nevada
Reno

# Human Body Parts Tracking: Applications to Activity Recognition

A thesis submitted in partial fulfillment of the requirements for the degree of Master of Science in Computer Science and Engineering

by

Aras Dargazany

Dr. Mircea Nicolescu, Thesis Advisor

December, 2011

University of Nevada, Reno
Statewide • Worldwide

## THE GRADUATE SCHOOL

We recommend that the thesis
prepared under our supervision by

**Aras Dargazany**

Entitled

**Human Body Parts Tracking: Applications to Activity Recognition**

be accepted in partial fulfillment of the
requirements for the degree of

**MASTER OF SCIENCE**

Mircea Nicolescu, Ph.D., Advisor

George Bebis, Ph.D., Committee Member

Yantao Shen, Ph.D., Graduate School Representative

Marsha H. Read, Ph.D., Associate Dean, Graduate School

December, 2011

# Abstract

As cameras and computers became popular, the applications of computer vision techniques attracted attention enormously. One of the most important applications in the computer vision community is human activity recognition. In order to recognize human activities, we propose a human body parts tracking system that tracks human body parts such as head, torso, arms and legs in order to perform activity recognition tasks in real time.

This thesis presents a real-time human body parts tracking system (i.e. HBPT) from video sequences. Our body parts model is mostly represented by body components such as legs, head, torso and arms. The body components are modeled using torso location and size which are obtained by a torso tracking method in each frame. In order to track the torso, we are using a blob tracking module to find the approximate location and size of the torso in each frame. By tracking the torso, we will be able to track other body parts based on their location with respect to the torso on the detected silhouette. In the proposed method for human body part tracking, we are also using a refining module to improve the detected silhouette by refining the foreground mask (i.e. obtained by background subtraction) in order to detect the body parts with respect to torso location and size. Having found the torso size and location, the region of each human body part on the silhouette will be modeled by a 2D-Gaussian blob in each frame in order to show its location, size and pose. The proposed approach described in this thesis tracks accurately the body parts in different illumination conditions and in the presence of partial occlusions. The proposed approach is applied to activity recognition tasks such as approaching an object, carrying an object and opening a box or suitcase. This approach shows promising results to future work that would result in a human body parts tracking system for recognizing more complicated activities which can lead to higher-level applications such as intent recognition.

# Dedication

This work is dedicated to my family for their help. I am thankful to them for their love and support.

# Acknowledgments

I would like to thank my advisor Dr. Mircea Nicolescu for his advice and encouragement.

Thanks to Dr. George Bebis and Dr. Yantao Shen for accepting to serve on my thesis committee.

I would also like to thank all my colleagues in Computer Vision Lab.

# Table of Contents

# List of Figures

# Chapter 1. Motivation and Goals

Human body parts' tracking has received a lot of attention in computer vision. The main reason is the necessity for recognizing human activities which is one of the most important goals in the high- level computer vision.

This project is intended for a surveillance application in a sensitive site, where it is desired to detect activities such as a person tampering with an object (e.g. opening a box).

In the first step, in order to know who is entering the site, we need to identify the person. There are several ways for people identification such as gait recognition and face recognition. Gait recognition is still not reliable enough due to low accuracy of recognition but face recognition is more reliable since face recognition only requires good face detection. In this context, we need to track the head in each frame until we are able to detect the face for face recognition purposes. In the second step, we need to find out whether the person is approaching the box or not. In order to recognize the approaching activity, we need to know the depth and the location of hand. In the third step, we like to see if the person in the room is opening the box to look at its content. Opening the box is difficult since we need to know first that the person approached the box and then box is being opened. For recognizing the box being opened, we need to detect the location of torso and hands. Last step is to detect if the person is carrying the box or not. For this activity, we need to know the location of hand, torso and object at each single frame from the time the person approached the box. In all these steps, we need to track human body parts in order to know the location of head, torso and hand at each frame to be able to recognize activities such entering the room, approaching the box, opening the box and carrying the box.

# Chapter 2. Related Work

There had been several applications for human body part tracking. The segmentation and tracking of the human body parts captivates the attention of the research community in computer vision due to their applications in automatic human activity understanding, gesture recognition, robotics or industrial control. Tracking systems that use video sequences captured with one camera make use of the color information [19], complex geometry of the body [11] or the dynamics of human motion [12]. Such systems often require a constraint background (i.e. constant background), a priori knowledge of the scale information, and human intervention in the initialization process. Human body parts tracking systems from multiple cameras [16,17,18] have demonstrated their improved performance in dealing with variations in scale and body occlusions. However, by their nature these methods are restricted to environments where a complex system with multiple cameras is available. Recently, consumer-level stereo cameras are becoming more commonplace and the performance of personal computers is approaching the threshold where stereo computation can be done at reasonable frame rates. Currently Kinect is introduced to the world; this stereo camera is produced and sold by Microsoft Inc. By running Kinect on 2.1GHz Pentium 4, 640x480 disparity maps are computed at 30 frames per second. The use of dense disparity maps together with color increases considerably the robustness of the system to variations in illumination conditions. It also reduces the inherent depth ambiguity present in 2D images and therefore enables accurate segmentation under partial occlusions and self-occlusions. As a result, robust techniques based on the use of stereo images and the depth information have been already considered for various applications such as pointing [15], tracking [14], or static gesture recognition [13]. In [9], a Bayesian model for the human upper body parts is used for segmentation and tracking of hand gestures using dense disparity maps. Like other body parts tracking systems, the proposed system in [9] also requires a user guided initialization for upper body segmentation by using a set of assumptions for relative position of the user. There had been several other methods proposed for human body part tracking based on using different features. Some were based on geometric features such as distances of body parts and their periodic movement [1]. The problem with this method is that in case of partial occlusion certain body parts may not be detected so that this method will not be able to detect body parts in a correct order. Since the main application of human body parts tracking is intent recognition and activity recognition, this method may not be reliable enough in the case of partial occlusion caused by multiple person interaction. Some other methods were using intensity-based features such as template matching masks (i.e. autocorrelation) within the region in which each body part is recognized in previous frames [2]. In [2], a template matching method is proposed for exact body part tracking which is performed in the probable region of each body part movements. This method is only able to track body parts in an image with no illumination changes and no occlusion since autocorrelation is an intensity-based method and performs well enough when there is not much intensity variation between frames. In the case of illumination changes, matching may not be performed well. In the case of occlusion, template matching (autocorrelation) is actually impossible. This method was very sensitive to lighting conditions. Also, if the movement of a body part or body part displacement was large and out of the probable searching region, template matching which is based on autocorrelation cannot find the body part in the probable searching region. In the real-world these methods discussed above are not really applicable.

In this project, intensity-based features (i.e. pixels intensity variation used in background subtraction for detecting foreground), geometry-based features (e.g. convex points on contour of silhouette) and combination of both SIFT and SURF which are actually local descriptors [3] as distinctive invariant features are all used.

In this thesis, we describe the implemented approaches for human body parts tracking in Chapter II, provide an overview of the proposed HBPT in Chapter III, describe the proposed HBPT in detail in Chapter IV and present some applications of HBPT to activity recognition with the experimental results in Chapter V.

# Chapter 3. Our Implemented Approaches

## 3.1 Convex Points on Silhouette Contour

This approach is using the convex hull points on the extracted contour of the silhouette. Using this approach, it is possible to find the convex and concave points on the boundary of silhouette which is available based on contour extraction. This method had already been used in [1]. The way it is used in [1] is by trying to approximate the location of each body part on the boundary and modeling each body part using the interest points on the boundary in that location.

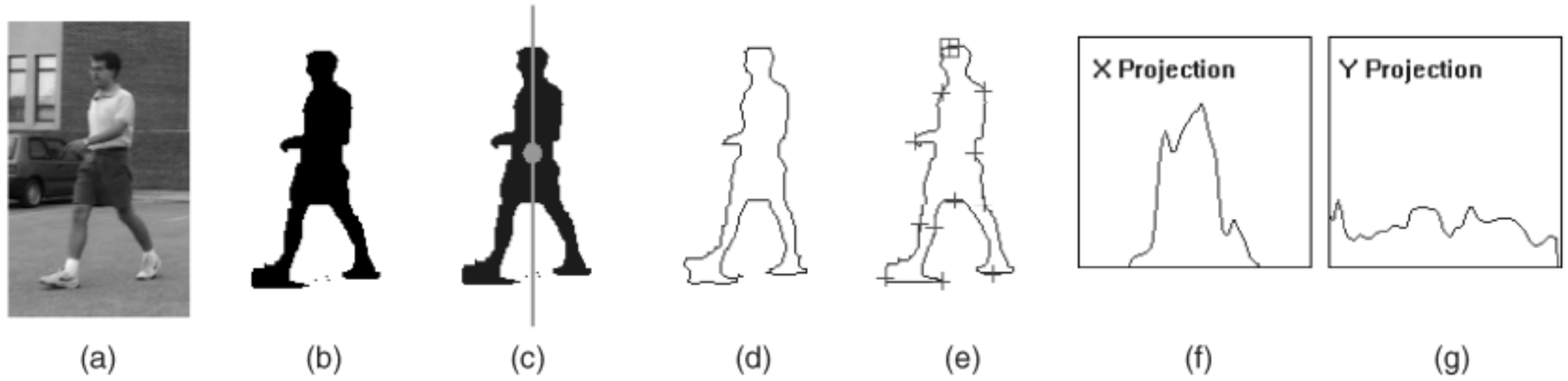

Fig. 3.1.1. Silhouette based shape features used in [1] : (a) input image, (b) detected foreground region, (c) its centroid and major axis, (d) contour of its boundary, (e) convex/concave hull vertices on its contour, (f) horizontal, and (g) vertical projection histogram

### 3.1.1 Foreground Detection and Silhouette Analysis

In this method, first we need to detect the silhouette. Once the silhouette was detected, we can determine the centroid of the silhouette using the bounding box center as shown in Foreground region using background subtraction as sown in figure 3.1.1.c. The shape of a 2D binary silhouette is represented by its projection histograms. Haritaoglu et al. [1] computes the 1D vertical (horizontal) projection histograms of the silhouettes in each frame. Vertical and horizontal projection histograms are computed by projecting the binary foreground region on an axis perpendicular to the major axis and along the major axis, respectively (Figs.3.1.1. f and g). Projection histograms are normalized by rescaling projections onto a fixed length, and aligning the median coordinates at the center.

Having found the silhouette and its centroid, we can obtain the boundary of silhouette by extracting its contour. We are using [8] as our contour extraction method.

Finally, this approach analyzes the shape of the silhouette boundary, already extracted in previous step using active contour, to find natural vertices as a candidate set of locations for body parts. We implemented two methods to find vertices on the silhouette boundary: a recursive convex hull algorithm (Graham scan) to find convex and concave hull vertices on the silhouette boundary and a corner detector in Harris et. al [20] based on local curvature of the silhouette boundary. The convex/concave hull algorithm gives better localization results but it is computationally expensive. Therefore, the corner detection algorithm is applied in every frame, but the convex/concave vertex algorithm is applied only when we need to detect the initial location of the body parts. Figs. 7d and 7e contains an example of a silhouette boundary and its corner points.

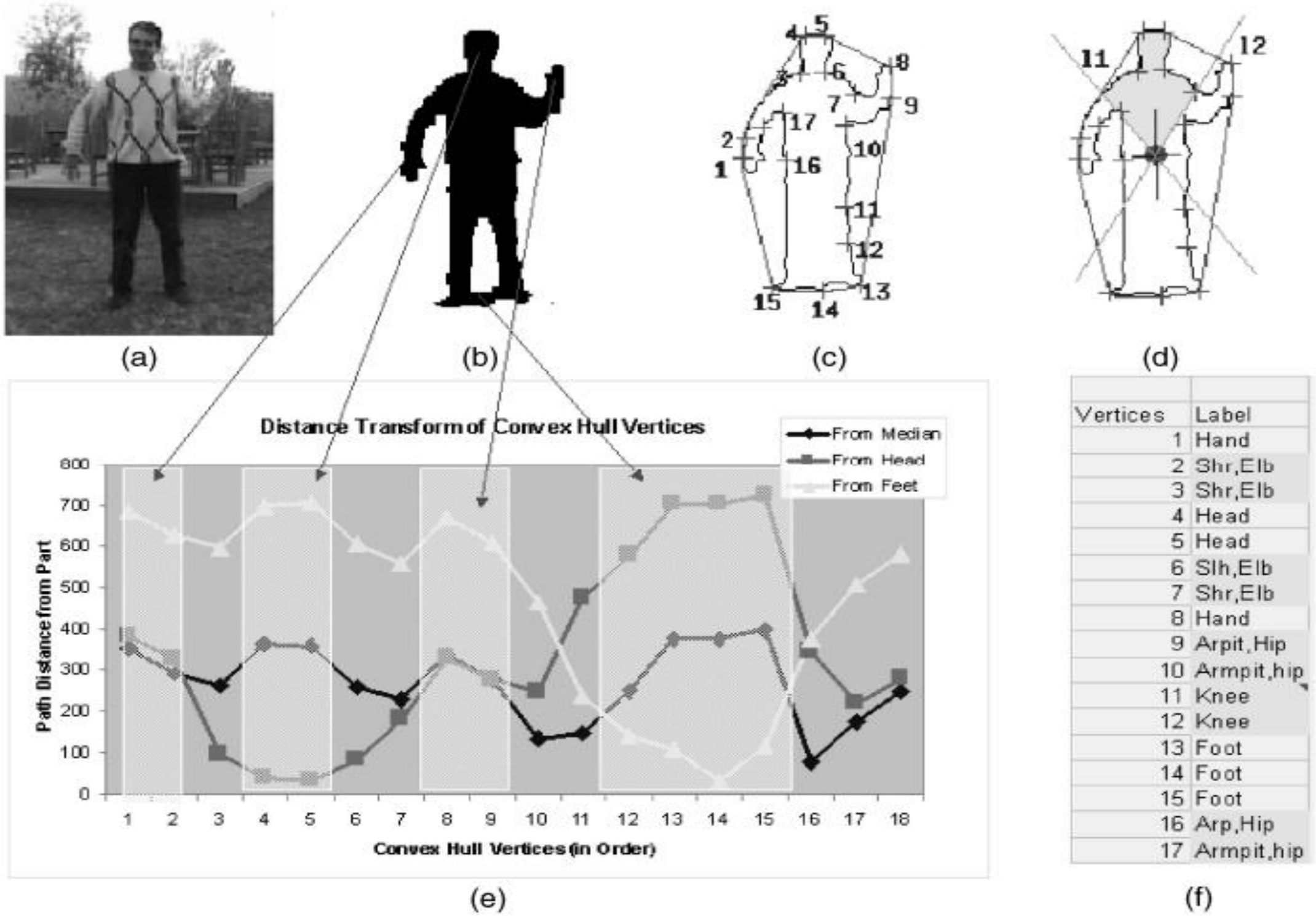

Fig. 3.1.2. An example showing how body parts are labeled in [1]: (a) original image, (b) detected silhouette, (c) detected convex and concave hull vertices, (d) silhouette segment for estimated head location, (e) distance transform measures for vertices from median, head and feet, and prelabeling of primary body parts (in shaded regions) after applying path distance constraints, (f) final labeling of body parts

### 3.1.2 Detection and Tracking of Body Parts Using Silhouette

Since detecting and tracking human body parts (head, hands, feet) is important in understanding human activities, we want to locate body parts such as the head, hands, torso and feet, and also track them in order to understand actions. The computational models employed by Haritaoglu et al.[1] to track the locations of the six main body parts (e.g., head, hands(2), feet(2), and torso) while a person is in any of a number of postures is shown in figure 3.1.2. The results of this method are shown in Fig. 3.1.3.

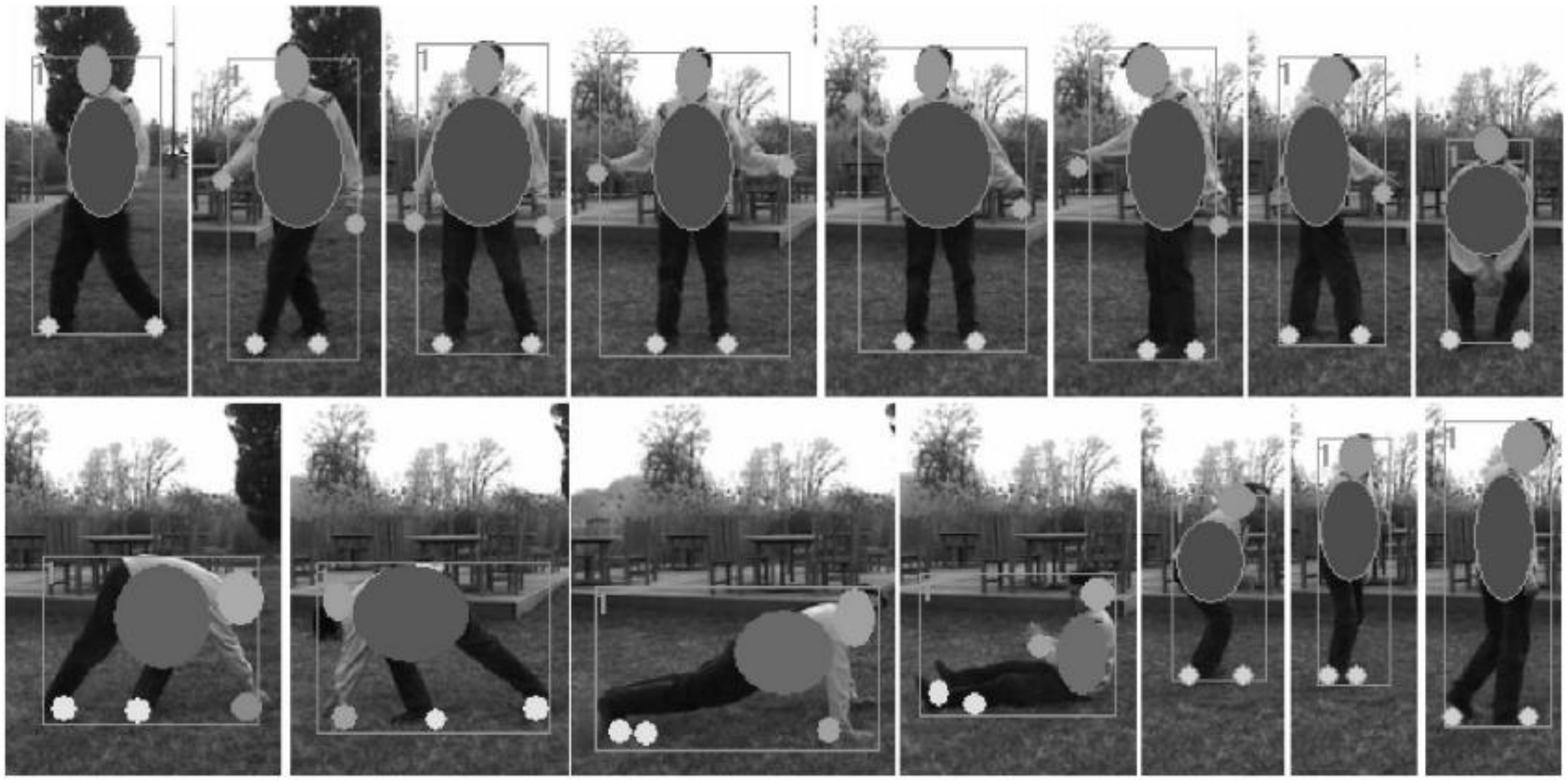

Fig. 3.1.3. Examples of using the silhouette model to locate the body parts in different actions in [1].

### 3.1.3 Implementation of this Method

For implementation of this method, first we apply background subtraction in order to detect foreground. Then we will find the contour of the detected silhouette. Once we found the boundary of silhouette, we determine the convex hull around the boundary. Using the convex hull, convex points are found on silhouette contour as shown in Fig. 3.1.4 for two consecutive frames in a sequence.

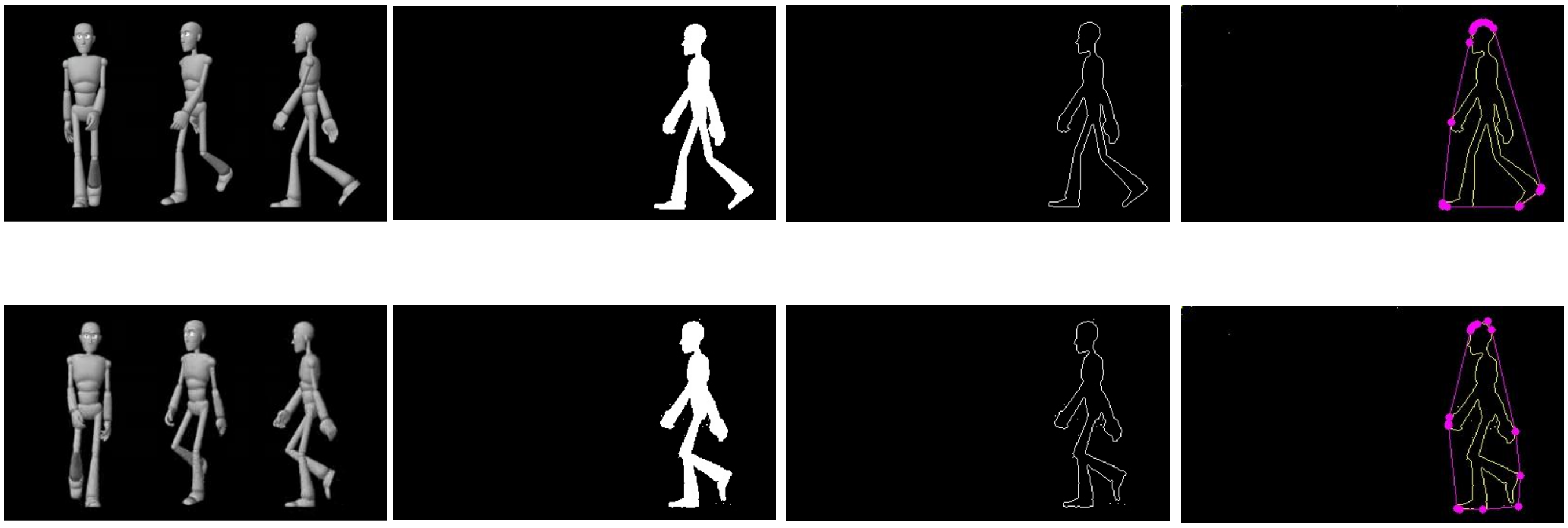

a b c d

Fig. 3.1.4. Silhouette analysis in our implementation: (a) input image, (b) detected foreground region, (c) contour of silhouette boundary, (d) convex hull vertices on silhouette contour

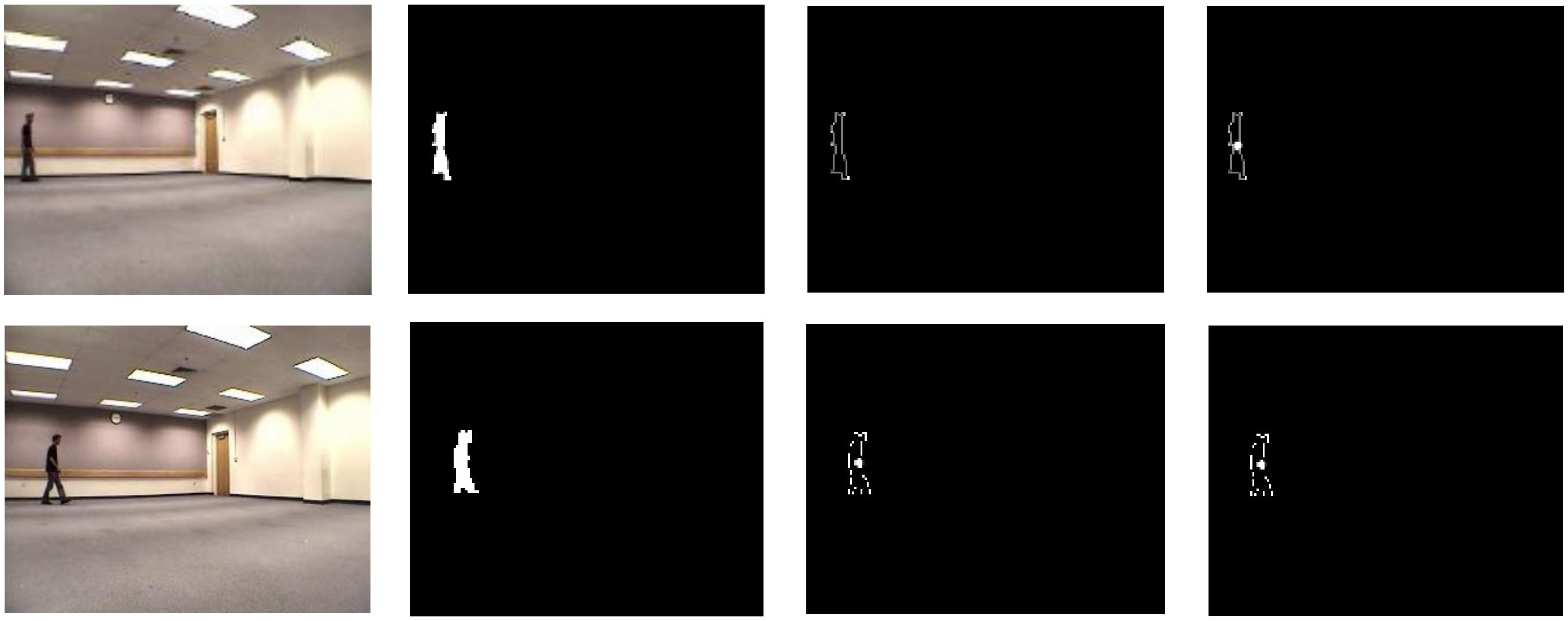

a b c d

Fig. 3.1.5. Silhouette analysis in real images: (a) input image, (b) detected foreground region, (c) contour of silhouette boundary, (d) centroid of the bounding box on silhouette contour.

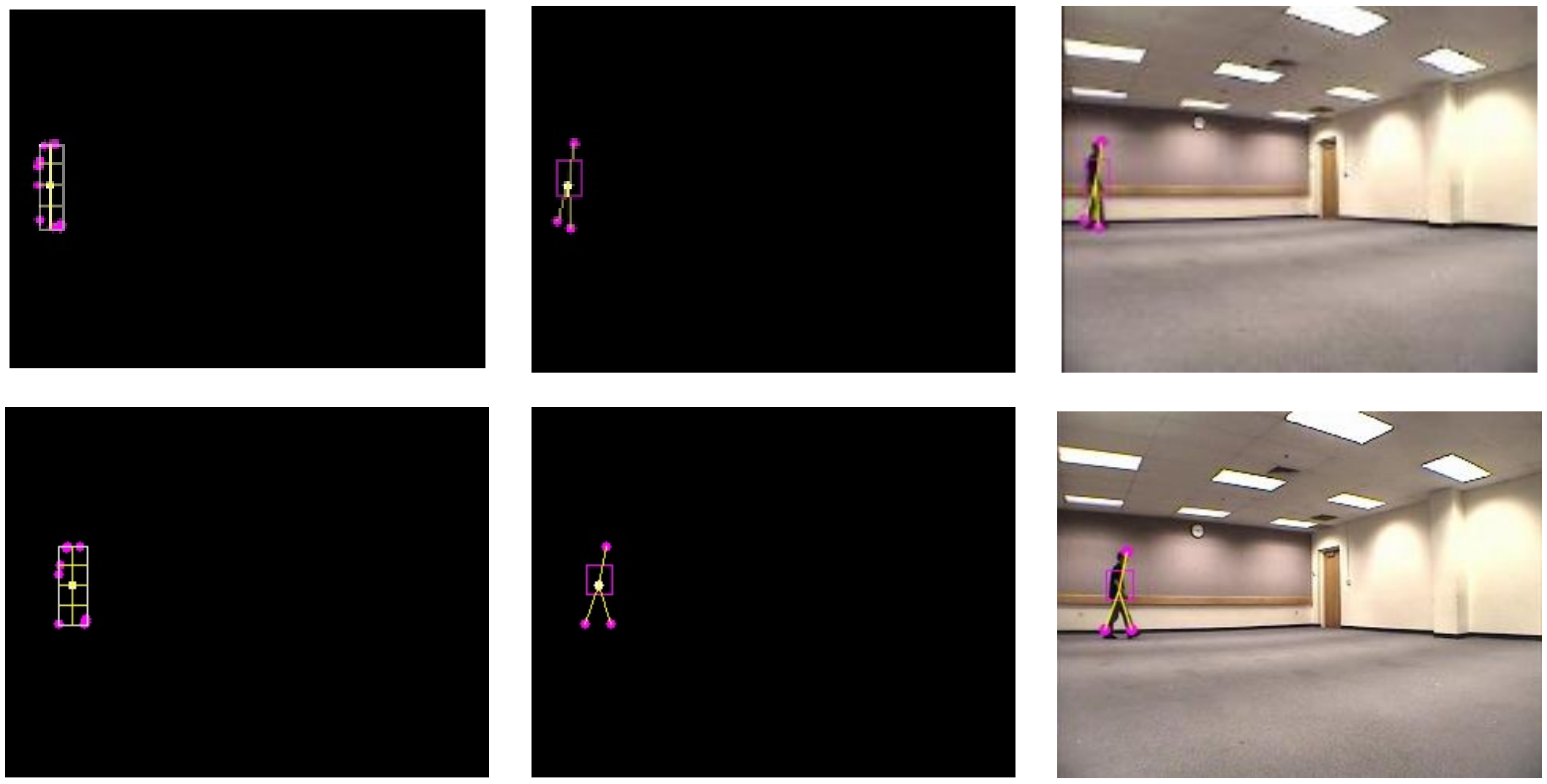

a b c

Fig. 3.1.6. Detection and tracking of body parts using Silhouette: (a) partitioning of convex points using their distance from the centroid, (b) the resulting body model for torso, head and feet, (c) body model superimposed on the real frame.

After we found the convex points, we will try to to do exactly the same on a real video to find convex points, centroid. According to the explained approach, we can build the body parts model based on the distances of the convex and concave points on boundary from the centroid. The final result of this implemented approach is shown In Fig. 3.1.5, 6, 7. In Fig. 3.1.5 , the results of this implemented method in real world show that the performance of this approach can be acceptable with a perfect silhouette. Fig. 3.1.6 shows how this method detects and tracks head, feet and torso using the distances of convex points from the centroid of the silhouette.

One of the important drawbacks of this method is that it requires good foreground detection in order to have perfect silhouette so that in the case of uneven illumination and occlusion, it may have serious problem in tracking body parts. The other problem is finding convex points on silhouette contour when the silhouette is not connected. This means if the silhouette is not connected, the contour will not be connected either and then there may be a lot of irrelevant convex points on the boundary of silhouette. Also detecting arms and hands is in this method, especially when the person is far from the camera as shown in the above figures. According to the results, this body parts tracking method based on the features on the boundary of silhouette (i.e. extracted contour) is not very reliable.

### 3.2 Distinctive Invariant Features and Transformation Function

This approach transforms image data (i.e. each frame data) into scale-invariant coordinates relative to local features. An important aspect of this approach is generating large numbers of features that densely cover the image (i.e. each frame) over the full range of scales and locations, specifically on silhouette. The quantity of features already extracted from silhouette is particularly important for human body part recognition in each frame. In order to detect each body part at least 3 points are required to be correctly matched from each body part in the current frame with each body part in previous frame for reliable identification and localization. The basic steps are as follows:

1) Learn the background in the first few frames (i.e. static camera).

2) Apply background segmentation to each frame in order to find silhouette (i.e. moving foreground) in each frame.

3) Extract keypoints and their descriptors for each frame using SIFT [3] or SURF [4]. Since SURF is faster than SIFT, SURF is a better choice in the case of body part tracking in a sequence of images.

4) Filter (i.e. keep) only keypoints on the silhouette using the detected foreground to initialize the tracker.

5) Specify manually a region around each body part (i.e. manual intervention)

using mouse with a rectangle.

6) Match distinctive invariant features (i.e. keypoints) on the silhouette in the current frame with features on the previous frame. This matching part is performed by a fast clustering method called FLANN (Fast Library for Approximate Nearest Neighbors) [5] which is currently the fastest clustering method. Using this clustering method, it is possible to find the keypoint with the closest descriptor (i.e. keypoint in the current frame or query frame) to the center of cluster (i.e. keypoint in previous frame or reference frame).

7) Find an affine transformation for each body part using at least three matches on the silhouette within the manually specified rectangle by determining its homography between the two sets of matched keypoints using RANSAC [6] to find the best matched keypoint candidates.

8) Localize each body part on the silhouette in the previous frame which will be performed by applying the affine transformation on four points of the rectangle (i.e. indicating the body part region) in the current frame.

In this approach, it is also possible to find the translation average between the matched keypoints on the detected silhouette (i.e. detected foreground using background subtraction) on the torso region which is initially specified by a user. This method will find the torso in the current frame by computing the average difference between the matched keypoints on the torso region in the previous frame and the current frame and then will apply the translation to the rectangle which is indicating the torso. The main problem is scaling: it is not really accurate to compute the scale changes of the torso by finding the average scale changes of the matched keypoints on the silhouette.

In order to find the keypoints on the silhouette, we need to initially have a constant background for background subtraction. Once we found the keypoints on the detected silhouette, we can initialize the system for tracking body parts in the frames with different background which is one of the significant contributions of this method compared to the others explained in this section. Below, results for this approach show the performance of this body part tracker for two frames in a sequence with different background. In all the results below, the background is not static.

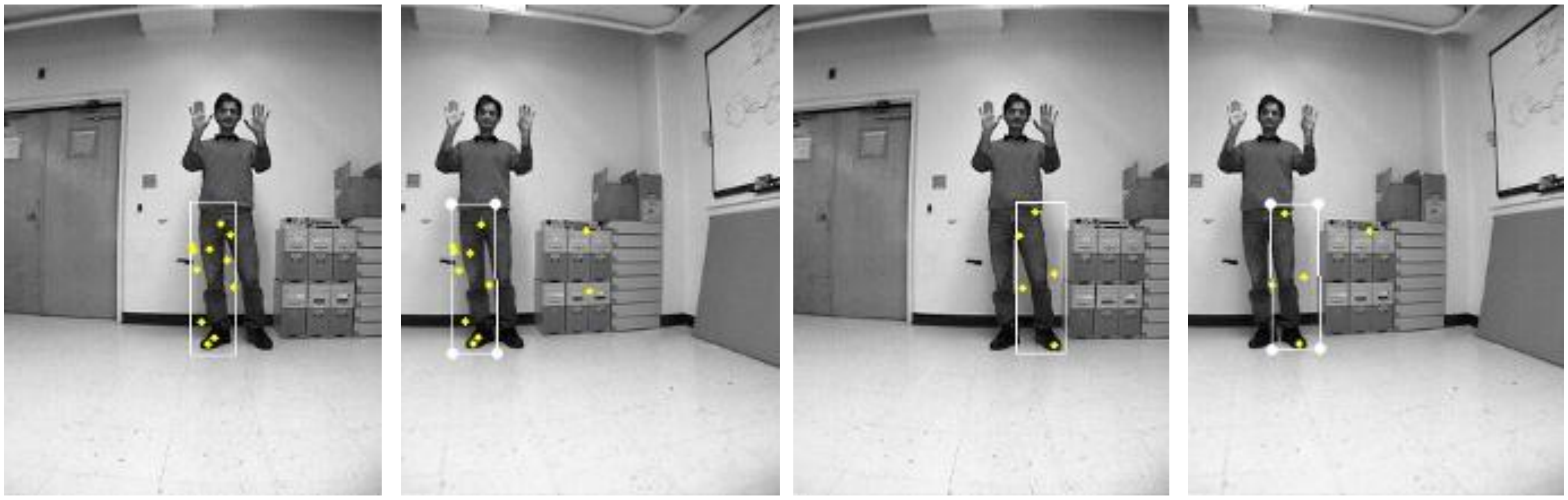

a b c d

Figure 3.2.1. Tracking of right leg and left leg. (a) previous frame with a specified region of right leg, (b) detected right leg with different background, (c) previous frame with a specified region of left leg, (d) detected left leg with different background.

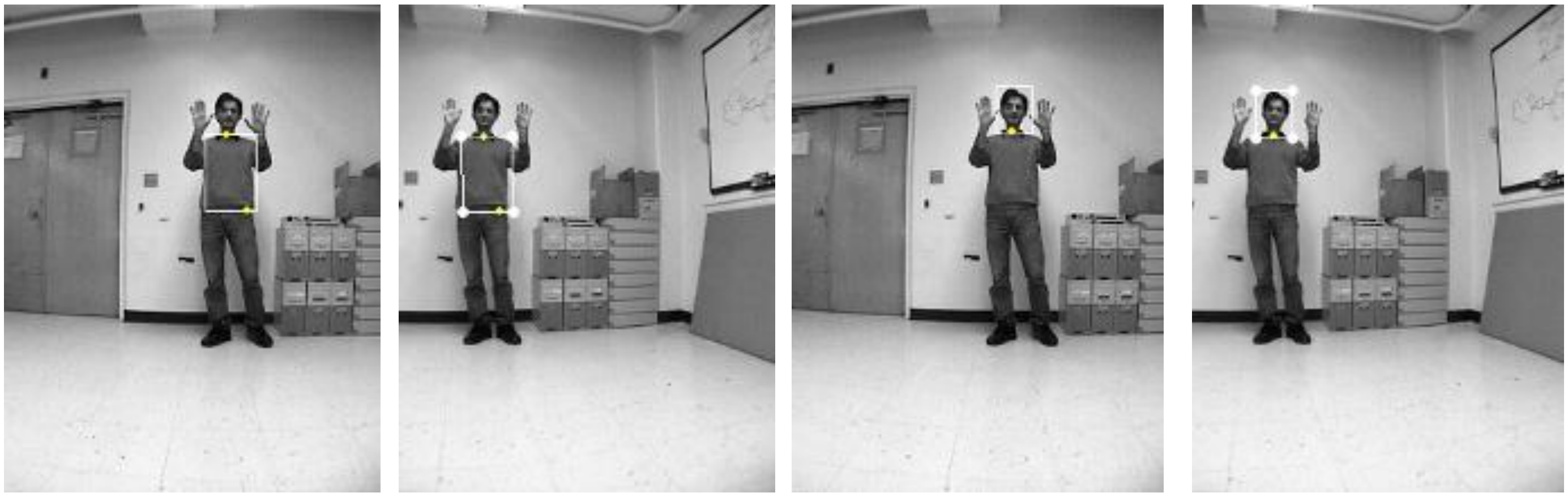

a b c d

Figure 3.2.2. Tracking of torso and head. (a) previous frame with a specified region of torso, (b) detected torso with different background, (c) previous frame with a specified region of head, (d) detected head with different background.

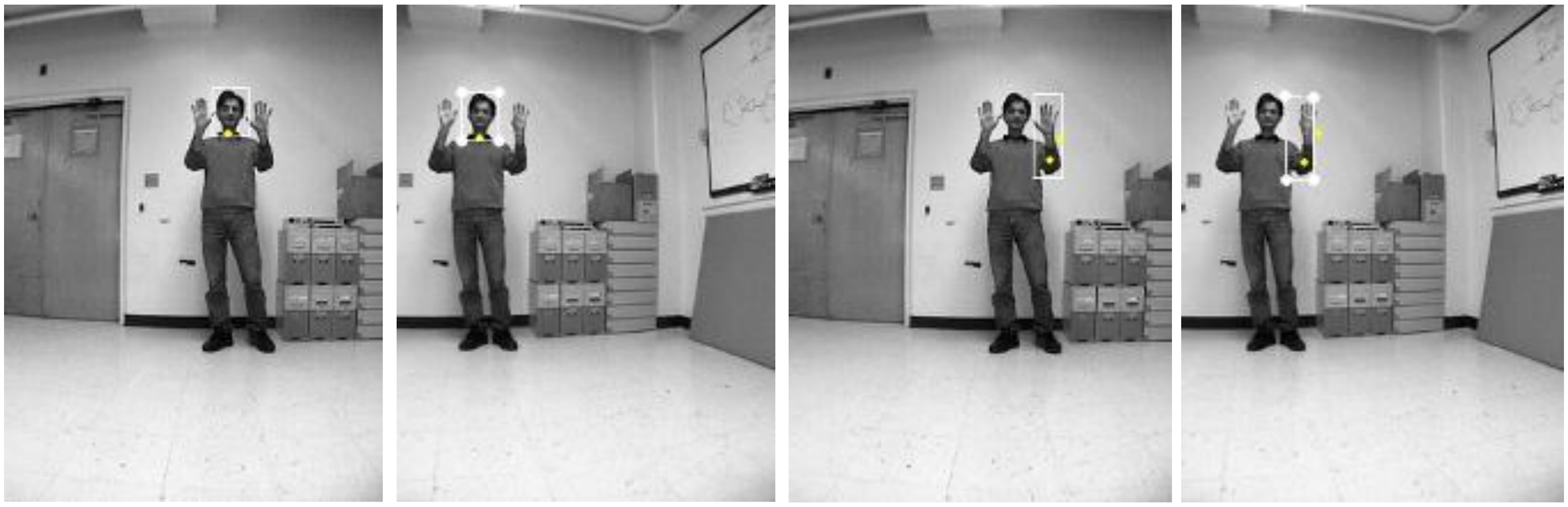

a b c d

Figure 3.2.3. Tracking of head and right hand as two close body parts: (a) previous frame with a specified region of face, (b) detected face with different background, (c) previous frame with a specified region of left hand which is really close to head, (d) detected left hand with different background and not getting distracted.

This method is performing well but it is not very accurate in a long term body part tracking. This body part tracking method is not giving promising results due to small displacement of keypoints in each frame and also non-rigidity of human body. This method may be performing really well for rigid objects such as a car, but for non-rigid objects like human body it is not very reliable. After a couple of frames, it is obvious that the resulting affine transformation from matched keypoints between previous frame and current frame on the silhouette cannot handle the scaling and translation of the keypoints between two consecutive frames. This problem will be even worse in the presence of scale changes.

# Chapter 4. The Proposed Approach Overview

In this section, we introduce the human body parts model and we give an overview of the proposed HBPT.

## 4.1. Modeling the Person

We can represent each human body part by 2D regions with their low-order statistics. Clusters of 2D points have 2D spatial means and covariance matrices. The blob spatial statistics are described in terms of their second-order properties; for computational convenience we will interpret this as a Gaussian model:

$$\Pr(O) = \frac{exp\left[-\frac{1}{2}(O-\mu)^T K^{-1}(O-\mu)\right]}{(2\pi)^{\frac{m}{2}}\,|K|^{\frac{1}{2}}} \tag{4-1}$$

Using this Gaussian interpretation, we can model each connected component (i.e. each body part) by a Gaussian blob since each connected component is in fact clusters of 2D points with spatial location indicated by **O**. The average of these 2D point spatial locations will give us the Gaussian blob spatial mean indicated. To compute the covariance matrix (**K**), we will use eigen-space analysis of a cluster for eigenvalue and eigenvector representations of each connected components as proposed in [21,24].

The Gaussian interpretation is significantly helpful to fit an ellipse to each body part (i.e. finding the appropriate blob for each body part region). The blob models represents the segmentation of the image into spatio-color classes. On the other hand, it means showing human body parts using ellipses or blobs as shown in figure 4.1.1.a. Like other representations used in computer vision, including modal analysis and eigenvector representations, blobs (i.e. ellipses) represent the global aspects of the shape.

Each blob has a spatial (x, y) and color (Y, U, V) component. Color is expressed in the YUV color space. Currently HBPT is restricted to use position and color only. Each blob can also have a detailed representation of its shape and appearance, modeled as differences from the underlying blob statistics. The ability to efficiently compute compact representations of people's appearance is useful for many applications [4].

Each blob is representing one human body part and as shown in figure 4.1.1.a, our human body parts model is composed of at most eight blobs representing head, torso, arm and legs. These blobs location, color, shape and appearance are updated in each frame to contain information about the location, size and pose of human body parts.

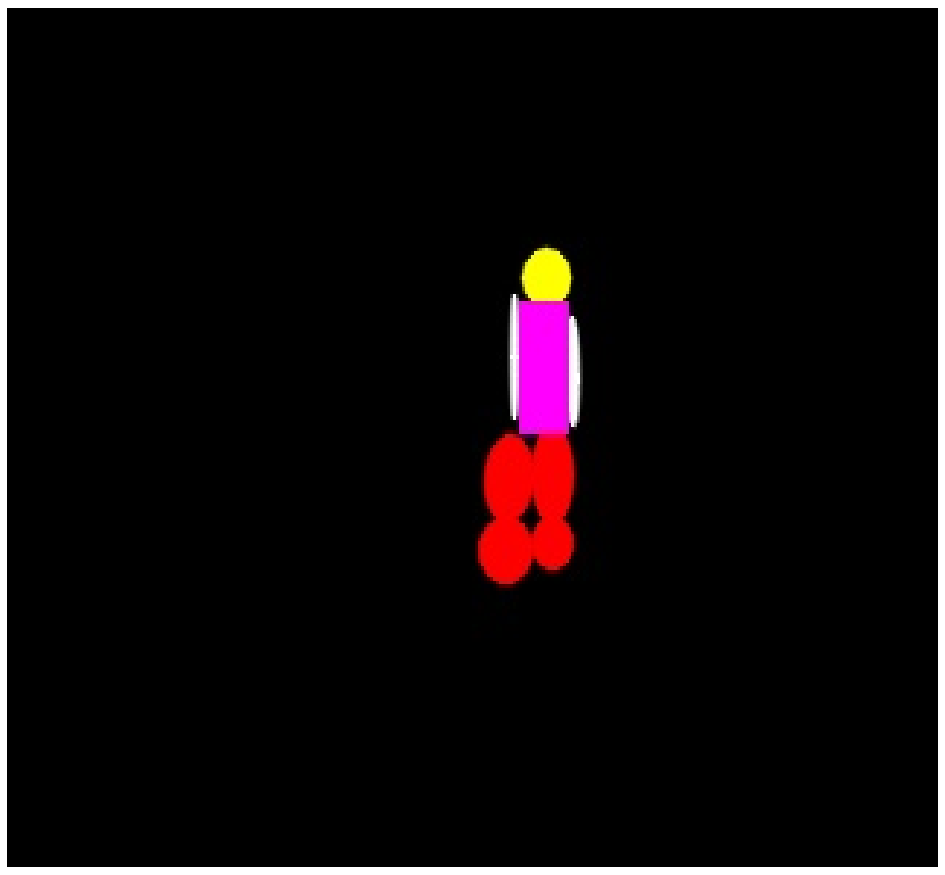

(a)

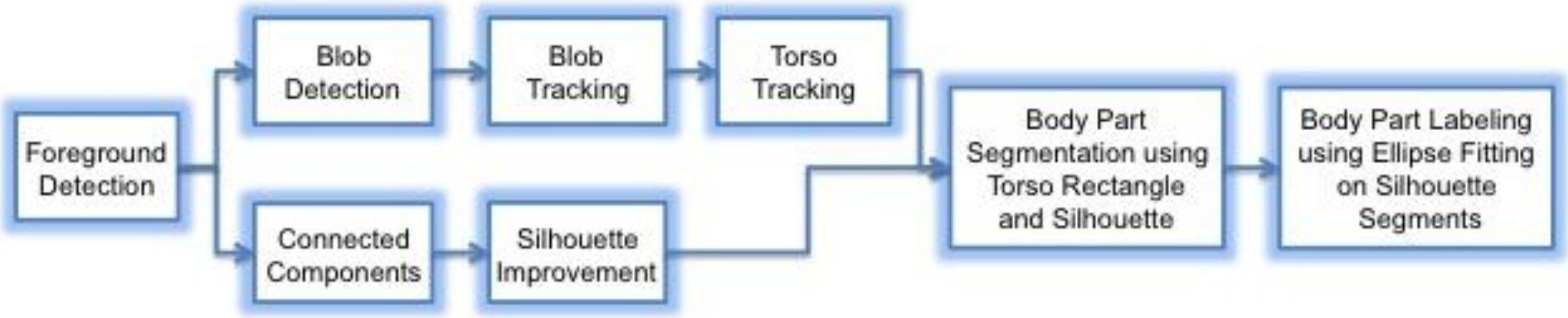

(b)

Figure 4.1.1.System overview, (a) 2D representation of human body parts model, (b) flowchart of proposed HBPT system

## 4.2. Modeling the Scene

We assume that the most of the time HBPT will be processing a scene that consists of a relatively static background such as in an office, and a single moving person. Consequently, it is appropriate to use different types of models for the scene and for the person. One of the important advantages of the proposed HBPT, as shown in figure 4.1.1.b, is its ability to recognize which scene pixels are occluded by the human (i.e. foreground), and which are visible (i.e. background). This information is critical in learning the background which will helps in background subtraction and therefore foreground detection. This is an essential step for many computer vision applications. In each frame, the proposed HBPT can detect the foreground by subtracting each frame from the learned background.

This allows us to compensate for changes in lighting, and even for object movement. For instance, if a person moves a book it causes the background to change in both the locations where the book was, and where it now is. By tracking the person, we can know that these areas, although changed, are still part of the scene model and thus they won't be detected as foreground. This background learning of proposed HBPT means that even large changes in illumination can be substantially compensated within two or three seconds.

## 4.3. Auto-Initialization of the Proposed HBPT

The proposed HBPT initialization process consists primarily of building representations of the person and the surrounding scene. It first builds the scene model by observing the scene without people in it, and then when a human enters the scene it begins to build up a model of that person.

The person model is built by first detecting the person as a foreground in the scene, and then building up a multi-blob model of the person body parts. This model building process is driven by finding the location of the torso which is in fact the largest ellipse contained in the silhouette and also the center of human body parts model. These two major characteristics of torso on the detected person's body help us to model it by a colored blob. Separate blobs are required for the person's arms, head, torso and legs.

The process of building a blob-model is guided by a 2D contour shape analysis that recognizes silhouettes in which the body parts can be reliably labeled. For instance, when the user faces the camera and extends both arms (what we refer to as the "star fish" configuration) then we can reliably determine the location of the head, hands and torso.

When the user points or reaches at something, we can reliably determine this action by detecting the location of the torso, head and hands. The following section describes the blob-model building process of the proposed HBPT in greater detail. In the next section, we describe the system with complete details.

# Chapter 5. Description of Our Proposed Approach

The proposed approach is composed of five different steps as described below:

## 5.1. Learning the Scene and Detecting the Foreground

Before the system attempts to locate a person in a scene, it must learn the scene. To accomplish this, the proposed HBPT begins by acquiring a sequence of video frames that do not contain a person. Typically this sequence is relatively long, a second or more, in order to obtain a good estimate of the color covariance associated with each image pixel. We tried 30 first frames of a sequence to learn the background. This foreground detection module is shown in figure 5.2.a. For background subtraction and foreground detection, we tested two different methods as follows:

1. The first method we tried was building an adaptive background mixture model based on [25]. The results of this method were found to be too noisy.

2. The more useful method that we tested was detecting foreground in complex background [26] which was a lot more useful in our proposed HBPT compared to the first method.

## 5.2. Detecting the Person Blob

After the scene has been modeled (i.e. background was learned), the proposed HBPT watches for large deviations from this model. New pixel values are compared to the known scene. If a changed region in the image is found, then that region will be detected as the foreground which is the first step in the proposed HBPT system as shown in figure 5.2.b. If the detected foreground is of sufficient size to rule out unusual camera noise, then the proposed HBPT proceeds to analyze the region in more detail, and begins to detect the region as a blob which is the first step to build up a blob model of the detected person.

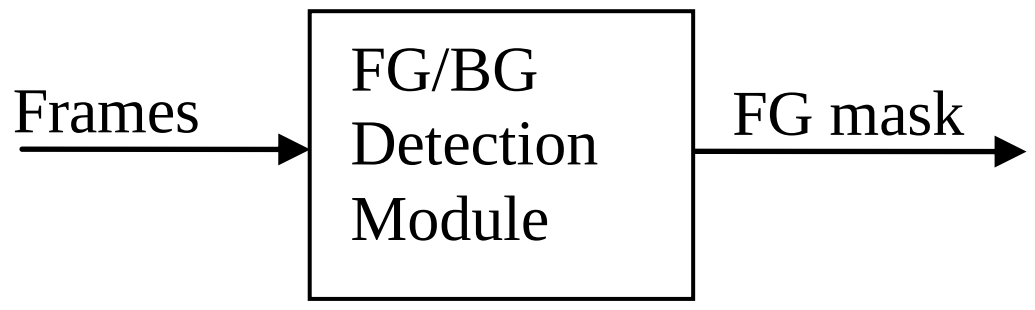

(a)

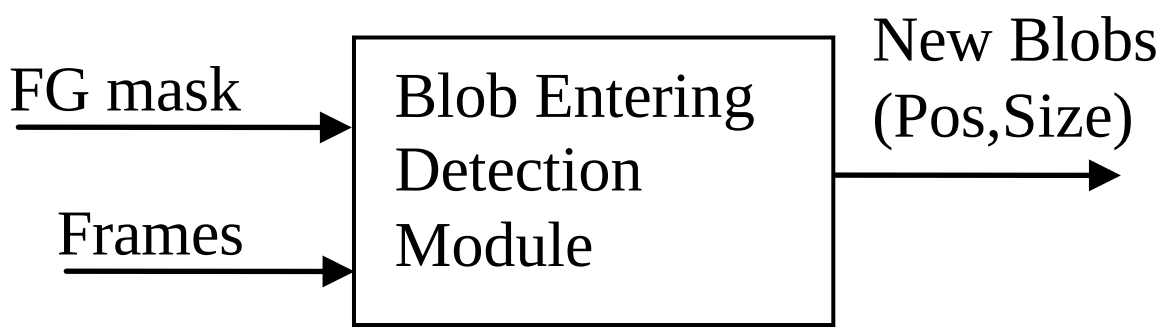

(b)

Fig. 5.2. The block diagrams of foreground detection module (a) and person blob detection module (b).

Once we detected the foreground, we need to detect the person. In this part we tried two different methods as shown below:

1. First we tested [27] which is actually based on detecting the person blob by tracking the connected components of the detected foreground. This method was not really successful in practice to detect the person.

2. Then we tried [27] to detect the person as a new blob by tracking the uniform moving of the connected components among the detected foreground. This was more useful in practice for the proposed HBPT since we know that we are finally interested in building the torso blob which is the largest ellipse contained within a connected component on the detected person silhouette.

### 5.3. Tracking the Person Blob

After the person blob is detected, we need to track the detected person blob in each frame as shown in figure 5.3.1. We tried five different methods as following:

1. Simple connected component tracking [23,27]. This one was not actually working

and was losing the target easily.

2. Mean Shift tracker [8,10,20]. This tracker was working but it was not accurate in tracking the target.
3. Applying Mean shift on the detected foreground [8]. The difference between this method and simple Mean Shift tracker is the initial point. In this method, we are using the detected foreground as the initial point for mean shift tracker. This means that the searching area for Mean Shift tracker will start on the detected foreground. This helps Mean Shift avoid false mode seeking which is usually the main problem with Mean Shift as a local mode seeking method. This problem is explained [8,10,20] and approached in [10]. The results of this tracker were almost the same as Mean Shift with slightly more accuracy.
4. Particle filter tracker optimized by Mean Shift [28]. This tracker performance was a lot better than previous methods. That is why we decided to try the combination of this tracker and connected components tracker [23,27] which was the next tracker we used.
5. Connected component tracking and MSPF [23,28] was performing well in the proposed HBPT in terms of tracking the detected person and tracking the torso of the detected person which was helpful to build the torso blob.

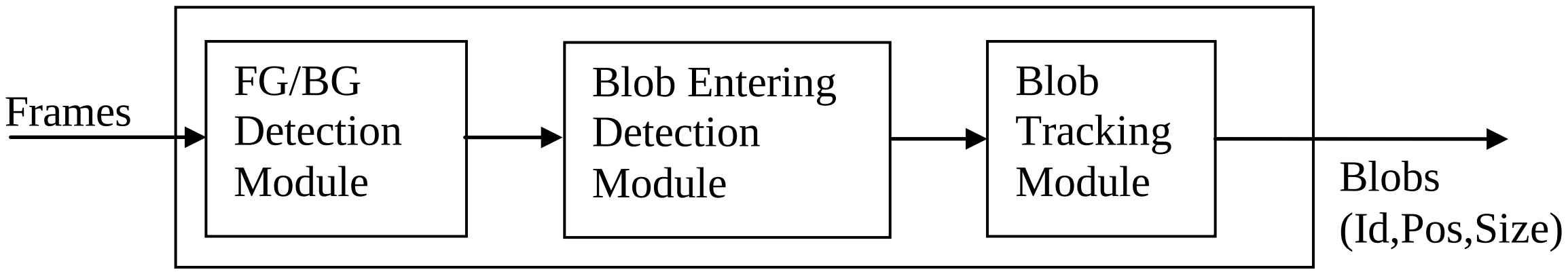

Fig. 5.3.1. The block diagrams of person blob tracking pipeline.

## 5.4. Improving the Foreground Mask

In order to build an acceptable blob model of human body parts, we need to enhance and improve the quality the detected foreground. Improving the quality of the detected foreground means to change the detected silhouette to one large connected component instead of many small connected components. To refine the detected silhouette, first we dilate, erode and dilate the foreground mask. Then we try to extract the contours of the foreground mask. To retrieve the contour of the connected components in the foreground mask, we are using a hierarchy method [10] which retrieves all of the contours and organizes them into a two-level hierarchy. At the top level, there are external boundaries of the connected components. At the second level, there are boundaries of the holes. If there is another contour inside a hole of a connected component, it is still put at the top level.

For contour approximation, we are using [6,11] which compresses horizontal, vertical, and diagonal segments and leaves only their end points. For example, an up-right rectangular contour is encoded with 4 points. Using this contour approximation, we can significantly improve the foreground mask and changed the number of connected components in the detected silhouette to almost one large connected component as shown in figure 5.4.1.

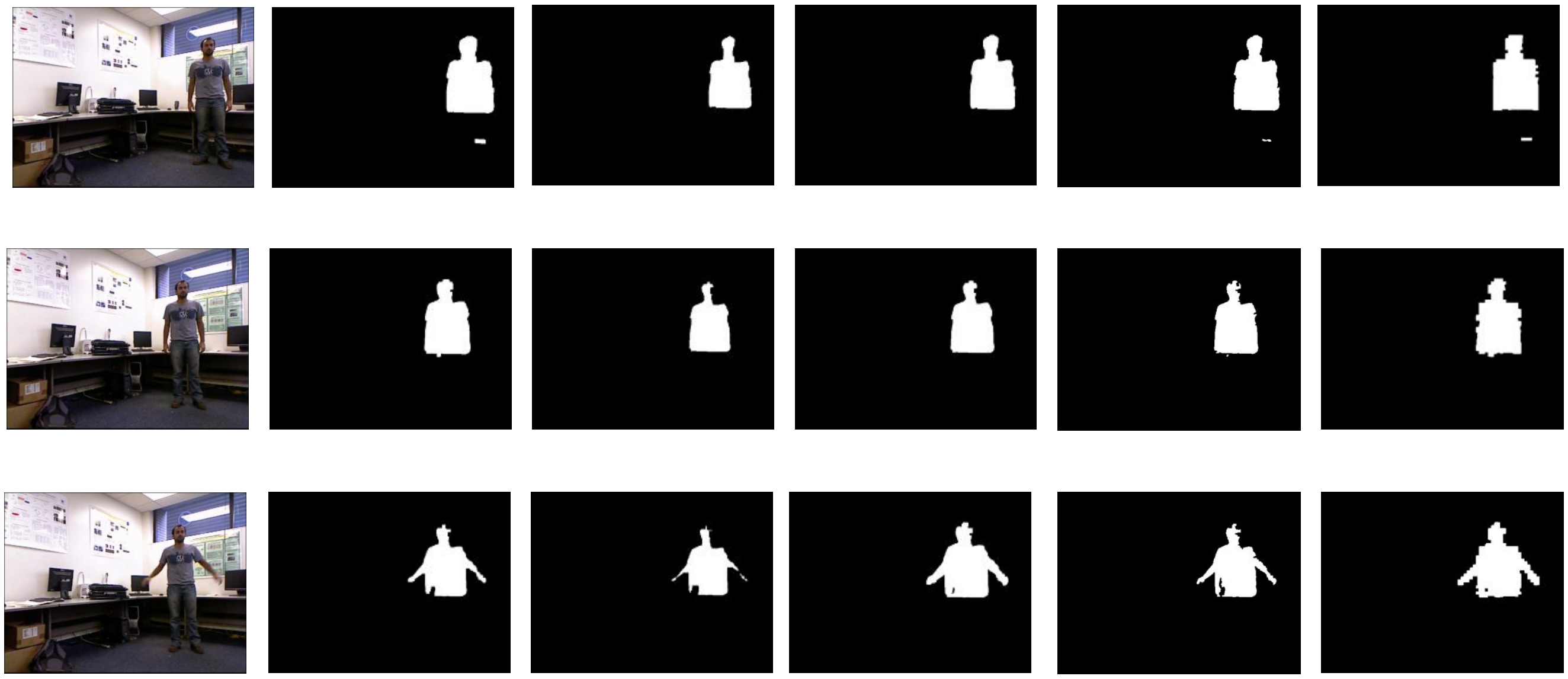

a b c d e f

Fig 5.4.1. The process of refining the foreground mask. (a) input frame, (b) dilated foreground mask, (c) eroding the dilated foreground mask, (d) applying dilation on the eroded and dilated foreground mask, (e) the initially detected foreground mask, (f) the final result of enhancing the foreground mask.

## 5.5. Tracking the Torso Blob

The results of tracking the person blob show that the center of this blob is always located on the torso region as shown in figure 5.4.1. Therefore we use the person blob location as the location of the torso blob. The main problem is the size since the person blob is larger than the torso blob. In fact, the person blob has a greater height than the torso blob and they both have almost the same width. In order to approach the problem of the torso blob size, we initialize the shape of the torso blob as a circle with the radius equal to the person blob width. In figure 5.5.1, the results of initialing the size of the torso blob using the person blob are also shown.

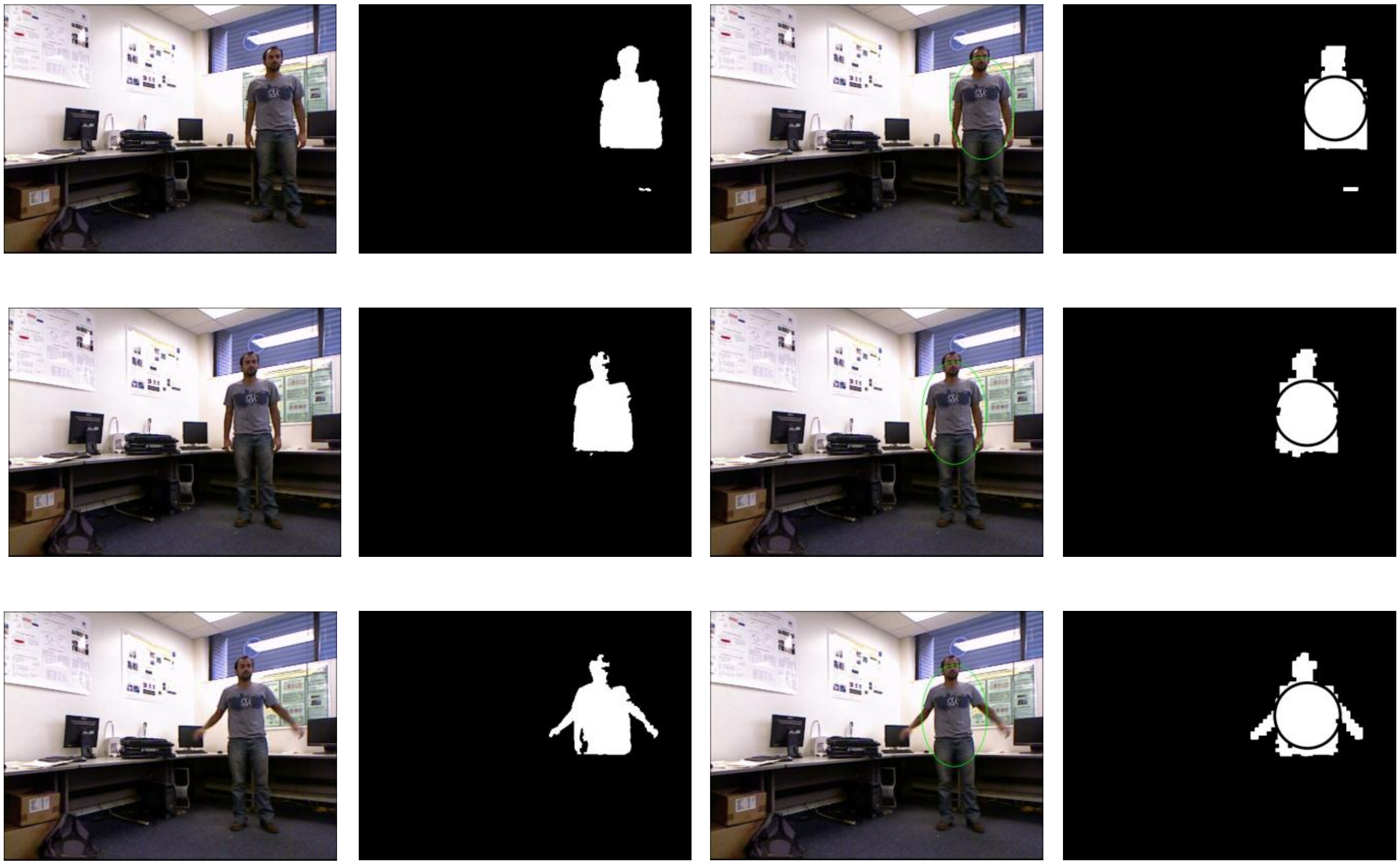

a b c d

Fig. 5.5.1. Tracking the torso blob using the person blob. (a) the input frame, (b) the detected foreground, (c) the person blob, (d) the torso blob on the improved detected foreground.

## 5.6. Building Blob Models of Body Parts

Blob models will be initialized based on the torso blob. We can divide the human body into five main regions with respect to the torso region, as follows:

- Central region inside the torso blob.
- Left arm region on the left side of the torso blob.
- Right arm region on the right side
- Head region above the torso blob
- Legs region below the torso blob

All these regions are disconnected from each other by the torso blob. To build blob models of body parts, the proposed HBPT uses a 2D contour shape analysis[10,11] on each single region to extract the related contour [22] of the silhouette in that region. Once the contour of silhouette in that region is retrieved, we model each of these contours by a Gaussian blob as discussed in section 4.1 for building a blob model to identify the head, arms, legs and torso locations.

This blob model identifies the location, size and pose of body parts. When a new blob for one body part is created, it is placed at its own region with respect to the torso blob (i.e. central region) such as at top, bottom and sides.

When there is no data to describe (as when an arm or head is occluded), the blob is deleted from the person model. When the arm or head later reappears, a new blob will be created and added to the torso blob. This deletion/addition process makes the proposed HBPT robust to occlusions and dark shadows.

The torso blob will be also updated since the retrieved contour in the central region might be model to a smaller torso blob than the initial torso blob which was used for dividing the regions on the silhouette. These details are illustrated in figure 5.6.1.

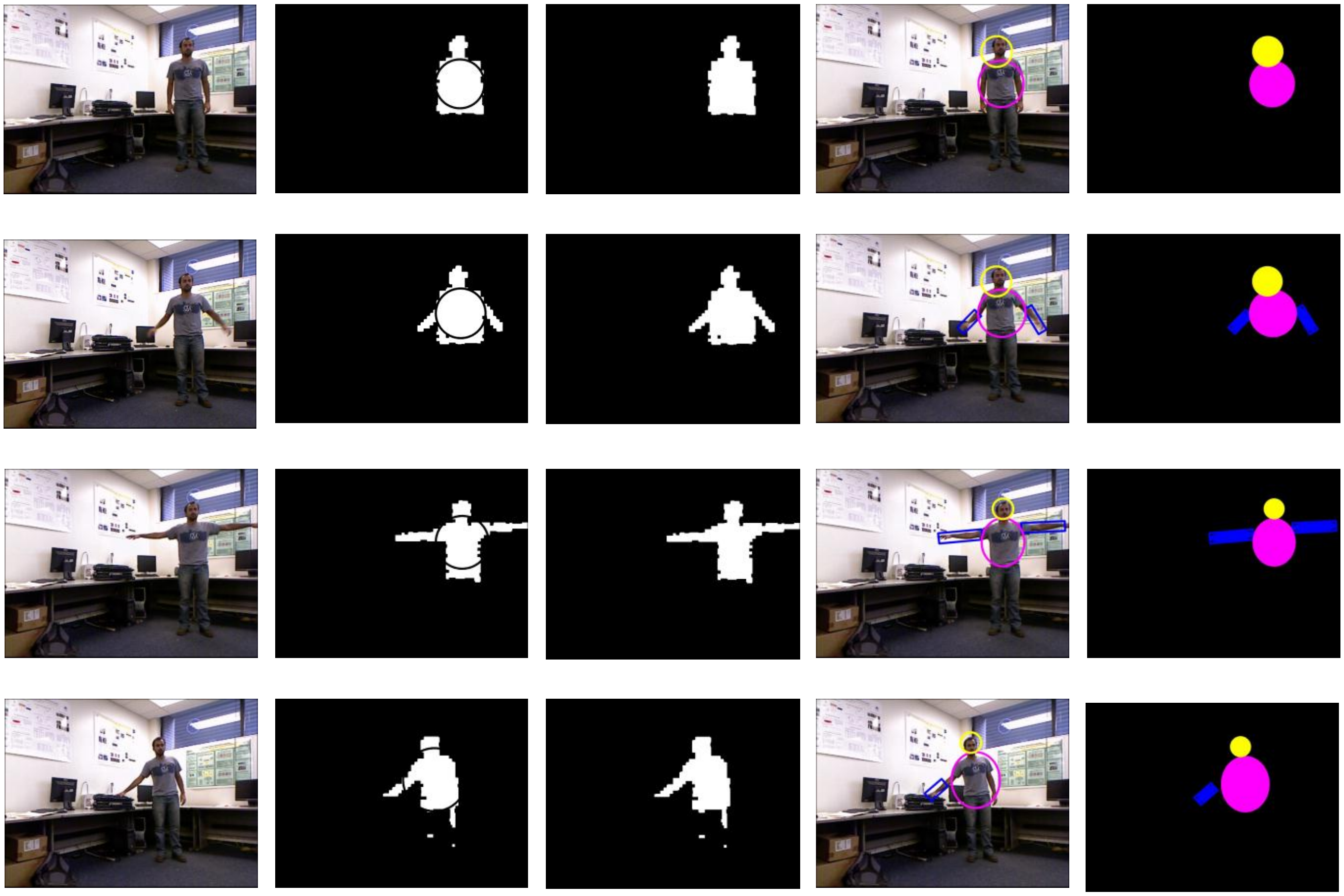

a b c d e

Fig 5.6.1. Building the blob model of human body parts by the proposed HBPT. (a) input frame, (b) applied initial torso blob on the refined foreground mask, (c) the refined foreground mask, (d) superimposing the blob model on input frame,

After building the blob models of body parts using the proposed HBPT if we also want to build a blob model for legs region, we need to have a bounding box around the refined detected silhouette in order to have an estimate of legs height. Once we draw a bounding box around the silhouette, we divide legs region (i.e. below the torso blob) into 4 sub-regions using the torso blob which gives us the width of the legs region, and height of the bounding box which will give the end of the legs. Using the torso blob and bounding box, we determine the width and the height of the legs region. Then we divide this region into four separate sub-regions as shown in Fig.5.6.2 and Fig.5.6.3.

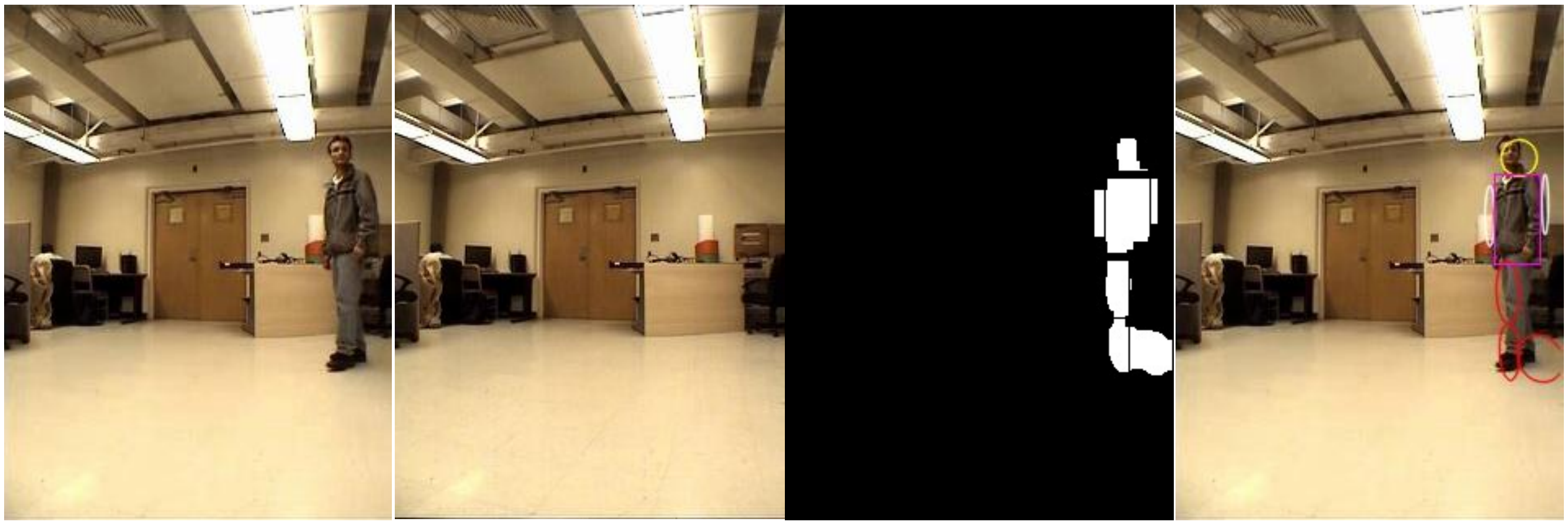

a b c d

Fig 5.6.2. The complete blob model of human body parts by the proposed HBPT. (a) input frame, (b) learned scene, (c) applied initial torso blob on the refined foreground mask and dividing the legs region by bounding box, (d) superimposing the blob model on input frame.

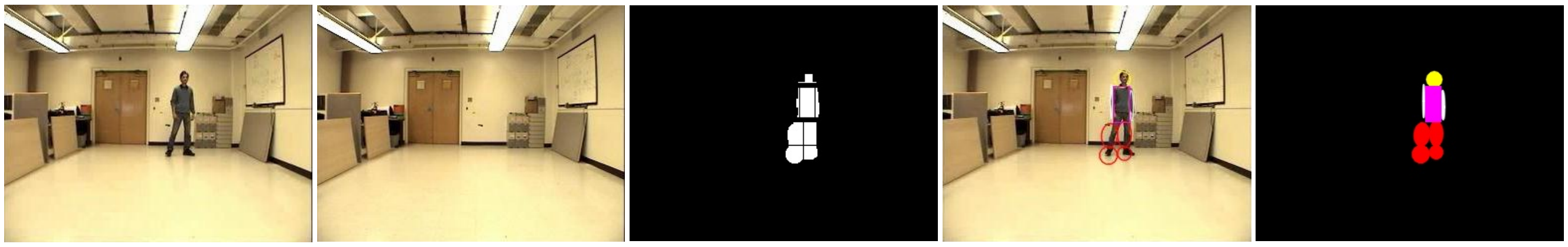

a b c d e

Fig 5.6.3. The complete blob model of human body parts by the proposed HBPT. (a) input frame, (b) learned scene, (c) applied initial torso blob on the refined foreground mask and dividing the legs region by bounding box , (d) superimposing the blob model on input frame (e) the resulting blob model of person body parts including legs.

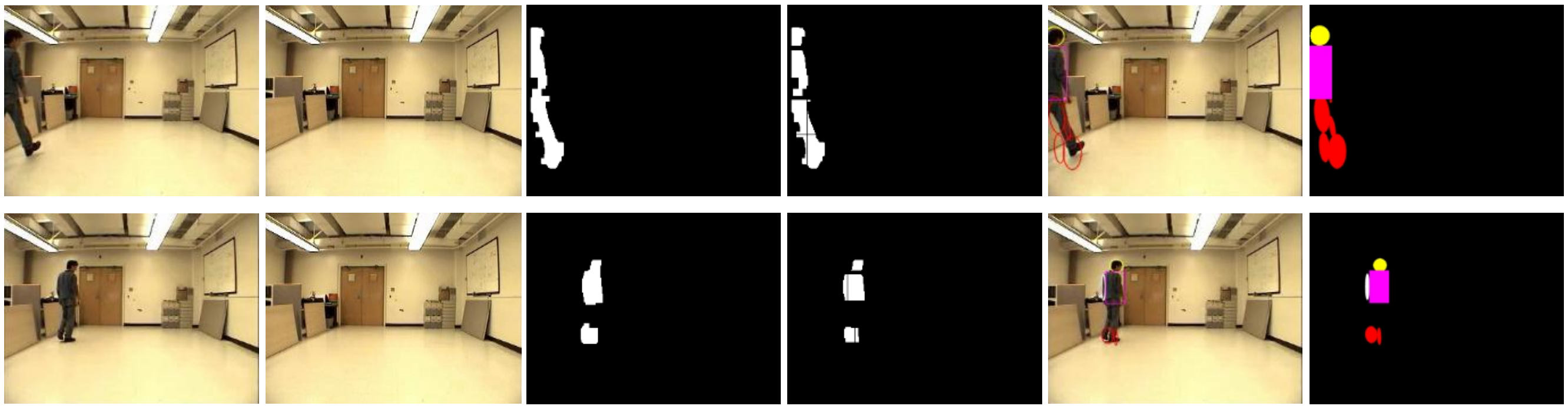

a b c d e f

Fig 5.6.4. The complete blob model of human body parts by the proposed HBPT. (a) input frame, (b) learned scene, (c) the refined foreground mask, (d) applied the torso blob on the refined foreground mask and dividing the legs region by bounding box, (e) superimposing the blob model on input frame (f) the resulting blob model of person body parts including legs.

In the next section, we describe how the proposed body part tracker can help in recognizing activities such as approaching a box, opening a box and carrying an object.

# Chapter 6. Applications and Experimental Results

In this section, we explain the application of the proposed HBPT in recognizing various surveillance-related activities:

## 6.1 Approaching a Box

We try to recognize an approaching box activity using the proposed HBPT. In order to be able to recognize such activity, we need to know the location of the hand with respect to the box.

1. We need to know the depth (i.e. z dimension) of hand and box in each frame along with the location of hand and box (i.e. x,y). This method is using the previously described technique to detect the foreground, detect and track the blobs in order to find the location of torso head and hands as shown in figure 6.1.2.

2. For depth recognition and measurement, we use Kinect device (i.e. Microsoft Kinect). Using Kinect, we are able to obtain the 3D point cloud of each frame which gives us the depth of almost every point in a frame as shown in figure 6.1.1. The Kinect can produce depth in a range of 10000 mm (10 meters), for a 640*480 frame. As shown in figure 6.1.1. (b) and (c), point cloud images of the same scene are not completely the same and there is a bit difference which will cause inconsistency in recognizing the depth.

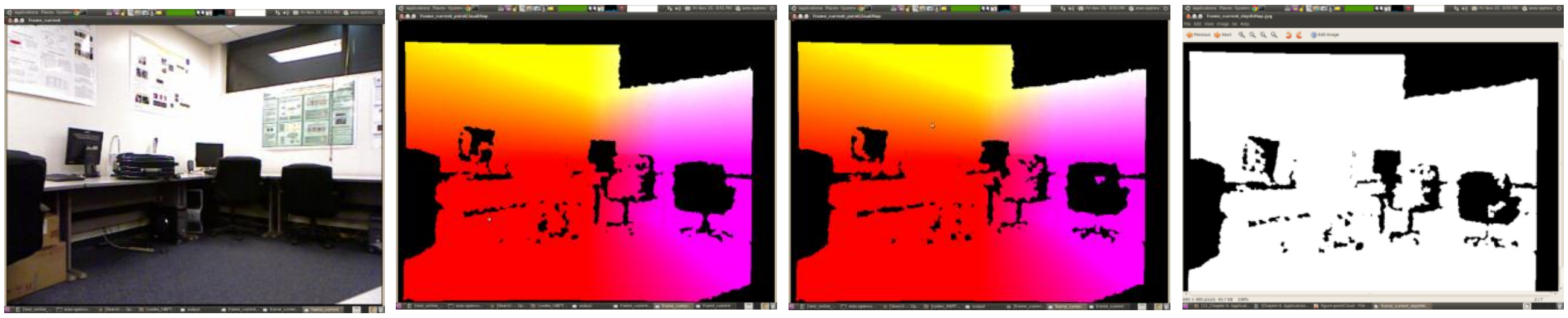

a b c d

Figure 6.1.1. Depth estimation using Kinect, (a) input frame, (b) point cloud in a colorized map, (c) point cloud in a colorized map after one second, (d) white area indicates locations in the image that have certain depth value.

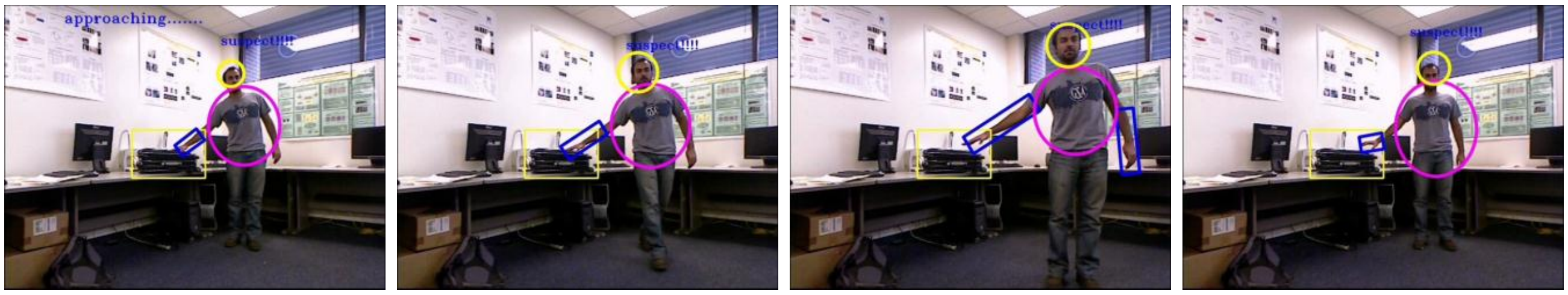

a b c d

Figure 6.1.2. Approaching box recognition, (a) successful detection of approaching box, (b) depth used when the person is not at the same depth as the box in order to avoid mistakes in box approaching recognition, (c) the person is standing far from the box based on depth data from Kinect, (d) the box and person are really close but still the box is not approached.

## 6.2. Opening a Box

Recognizing this activity is highly dependent on appearance changes. The pre-requisite step for this is approaching the box and initializing the system with the location and size of the box. As shown in figure 6.2.1 (a) and (c), the yellow rectangle is specified by user and it helps us build the reference histogram of the box inside the yellow rectangle. The red rectangle in figure 6.2.1 (b) is initialized by the yellow rectangle and it will move based on Mean Shift method [8, 10, 20] which is based on the difference between these two histograms that represent the color model of the area inside rectangles. If somebody approached the box, then we try to recognize the activity. We tried different approaches for recognizing this activity based on changes in the appearance of the box:

1. Detecting the opened box as foreground and then as a blob.

2. Building a color model of the rectangle area using a histogram with 16 bins and try to compare these histograms frame by frame to detect a very large difference using Mean Shift which will indicate opening box action.

3. The most robust approach was using Mean Shift [8] to find the difference of color histogram models of the box.

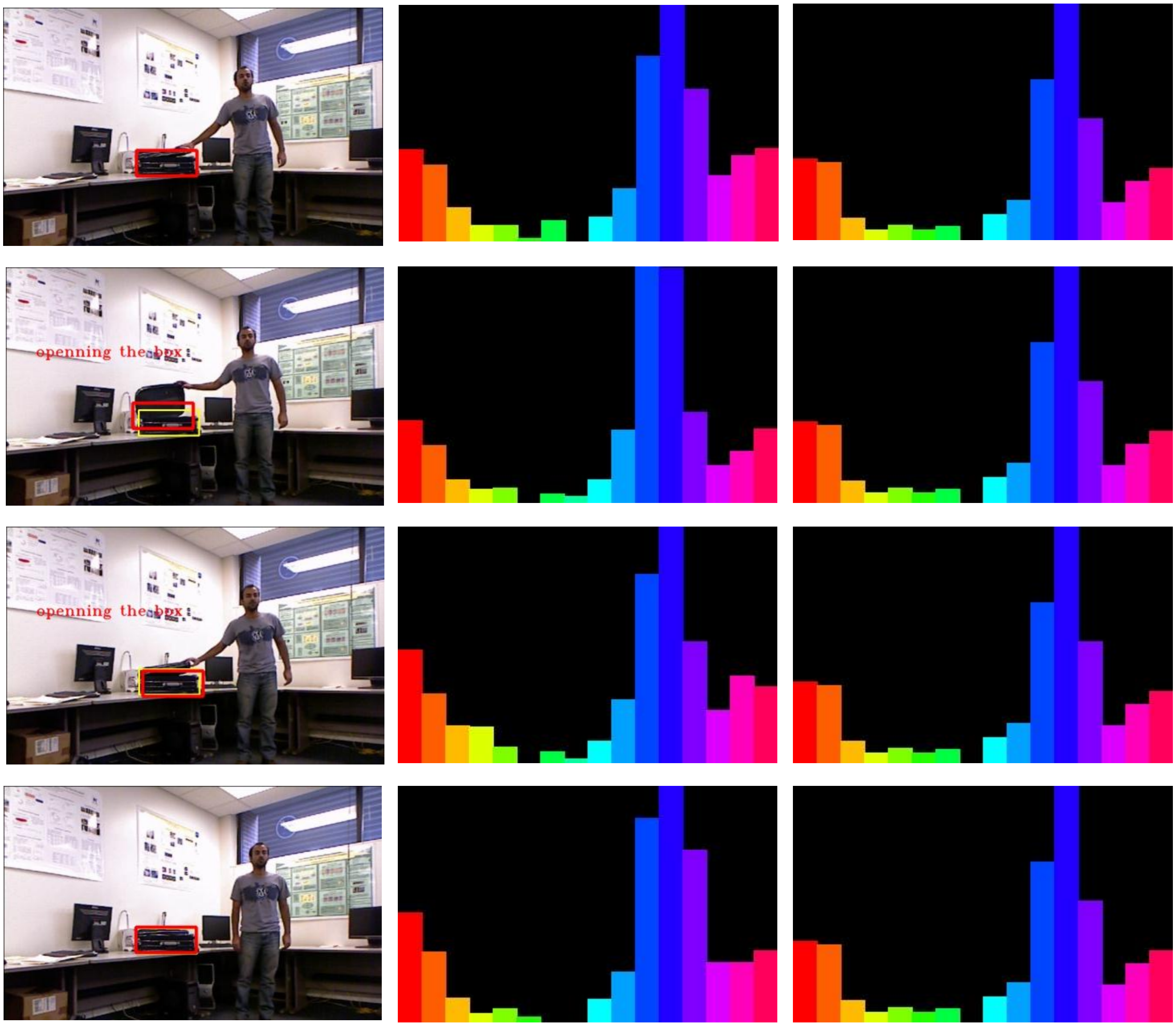

a b c

Figure 6.2.1. Opening box recognition, (a) computing the translation of red rectangle on yellow rectangle using mean shift, (b) color histogram of red rectangle, (c) color histogram of reference frame for yellow rectangle which gives reference histogram.

### 6.3. Carrying an object

In order to be able to recognize this activity, it is required to recognize an approaching activity first. Once we recognized the approaching activity, we determine whether someone is carrying the object or not. In order to recognize this activity, we obtain the location of the hand with respect to the object (i.e. already approached by the tracked person). To achieve this goal, we should have:

1. An object tracker to give the location of the object at each frame. We are using Lucas-Kanade tracking method which is based on optical-flow and it is known as a point tracker.
2. The depth information given to us by Kinect.
3. The location of the hand and torso which is obtained using the proposed HBPT.

Using this information, we are able to measure the distance between the hand and the object from the time the object is approached. We also compare the location of the object in each frame with the previous frame in order to make sure that the object is being carried as shown in below figures.

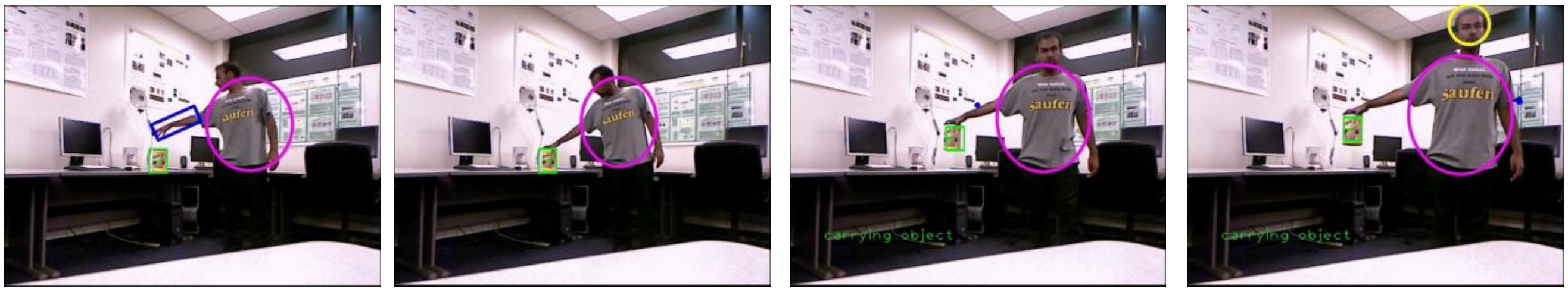

a b c d

Fig 6.3.1. Recognizing object carrying using the proposed HBPT, (a) getting close to the object, (b) approaching the object (c) recognition of object carrying by depth data from Kinect when the person and object are getting closer to camera, (d) the object and person are really close to camera.

# Chapter 7. Conclusions and Future Work

One of the main contributions of the proposed HBPT is that it does not require any initialization or manual intervention in terms of body parts tracking. Also it can be easily used in surveillance- related applications or it can be extended to other use, such as human-computer interaction.

The proposed HBPT has some drawbacks. One of the important problems with the proposed HBPT is that it needs constant background which is not really desirable for application with a dynamic background. This constant background is essential for background subtraction. Another problem with this method is that it works only close to real-time (10 frames per second) which needs to be improved in order to be applicable in real world. The other issue with the proposed HBPT is that this method is not perfectly robust to illumination changes, although the blob tracking method used in the proposed HBPT has the best performance among techniques we tried since it is performing almost real-time and even if it does not have a perfect silhouette, it can still perform well enough for the intended applications. The last problem with the proposed HBPT is not being able to track hands when they are inside the torso region. We next explain that we can approach this problem using depth value in each frame.

This approach is showing acceptable different body parts tracking accuracy with their pose even in the presence of highly perspective transformations on each body part, specifically hands. This method is applied to online-body part tracking as a close to real-time application as mentioned in the previous section.

Some extensions to improve this approach are:

• The proposed HBPT approach can be more robust if we incorporate depth to each frame (3D) using the Kinect sensor for better localization of body parts.

• Using the history of location and size of body parts in previous frames, we should be able to track the body more accurately and faster.

• We can extend this method to articulated motion of body parts in order to have more details about the movement of smaller body parts.

• In order to recognize activities such as running and walking, we will also need leg tracking.

# References

[1] I. Haritaoglu, D. Harwood, and L.S. Davis, “w4 : Real-Time Surveillance of People and Their Activities”, IEEE Trans. Pattern Analysis and Machine Intelligence, vol. 22, no. 8, pp. 809-830, Aug. 2000.

[2] J. Ben-Arie, Z. Wang, P. Pandit and S. Rajaram, “Human Activity Recognition Using Multidimensional Indexing ”, IEEE Trans. Pattern Analysis and Machine Intelligence, vol. 24, no. 8, pp. 1091-1104, Aug. 2002.

[3] D. G. Lowe , “Distinctive Image Features from Scale-Invariant Keypoints ”, International Journal of Computer Vision 60(2), 91–110, 2004.

[4] Bay, H., Tuytelaars, T., & Van Gool, L., “SURF: Speeded-Up Robust Features”, 9th European Conference on Computer Vision, Vol. 110, pp. 346-359, 2008.

[5] M. Muja and D. G. Lowe, “Fast Approximate Nearest Neighbors with Automatic Algorithm Configuration”, International Conference on Computer Vision Theory and Applications (VISAPP'09), 2009.

[6] M. A. Fischler and R. C. Bolles, “Random Sample Consensus: A Paradigm for Model Fitting with Applications to Image Analysis and Automated Cartography”, Comm. of the ACM 24: 381–395, June 1981.

[7] R. Lienhart, Jochen Maydt, “An Extended Set of Haar-like Features for Rapid Object Detection”, ICIP, 2002.

[8] D. Comaniciu, V. Ramesh, and P. Meer, “Kernel-based object tracking”, PAMI, 2003.

[9] R. D. Cavin, A. V. Nefian and N. Goel, “A Bayesian Formulation for 3D Articulated Upper Body Segmentation and Tracking from Dense Disparity Maps”, ICIP, 2003.

[10] A. Dargazany, A. Soleimani, A. Ahmadyfard, “Multi-bandwidth Kernel-Based Object Tracking”, Hindawi Publishing Corporation, Advances in Artificial Intelligence, Article ID 175603, 15 pages, 2010.

[11] C. Barron and I. Kakadiaris. Estimating anthropometry and pose from a single image. In Computer Vision and Pattern Recognition, pages 669–676, 2000.

[12] C. Bregler and J. Malik. Tracking people with twists and exponential maps. In Computer Vision and Pattern Recognition, pages 8–15, 1998.

[13] R. Grzeszczuk, G. Bradski, M.H. Chu, and J.Y. Bouguet. Stereo based gesture recognition invariant to 3D pose and lighting. In International Conference on Computer Vision and Pattern Recognition, pages 826–833, 2000.

[14] N. Jojic, B. Brumitt, B. Meyers, S. Harris, and T. Huang. Tracking self-occluding articulated objects in dense disparity maps. In International Conference on Computer Vision, pages 123–130, 1999.

[15] N. Jojic, B. Brumitt, B. Meyers, S. Harris, and T. Huang. Detection and estimation of pointing gestures in dense disparity maps. In International Conference on Face and Gesture Recognition, pages 468–475, 2000.

[16] I. Kakadiaris and D. Metaxas. Model-based estimation of 3D human motion. IEEE Transactions on Pattern Analysis and Machine Intelligence, 22(12):1453–1459, 2000.

[17] A. V. Nefian, R. Grzeszczuk, and V. Eruhimov. "A statistical upper body model for 3D static and dynamic gesture recognition from stereo sequences", In IEEE International Conference on Image Processing, pages 601–607, 2001.

[18] H. Sidenbladh, F. De La Torre, and M. J. Black. A framework for modeling the appearance of 3D articulated figures. In Automatic Face and Gesture Recognition, pages 368–375, 2000.

[19] C. Wren, A. Azerbayejani, T. Darell, and A. Pentland. Pfinder: Real-time tracking of the human body. IEEE Transactions on Pattern Analysis and Machine Intelligence, 19:780–785, July 1997.

[20] Aras Dargazany, Ali Soleimani, "Kernel-Based Hand Tracking", INSInet Publication, Australian Journal of Basic and Applied Sciences, 2009.

[21] A. Azarbayejani and A. Pentland, "Recursive Estimation of Motion, Structure, and Focal Length," Trans. Pattern Analysis and Machine Intelligence, vol. 17, no. 6, pp. 562–575, June 1995.

[22] A. Baumberg and D. Hogg, "An Efficient Method for Contour Tracking Using Active Shape Models," Proc. Workshop Motion of Nonrigid and Articulated Objects. Los Alamitos, Calif.: IEEE CS Press, 1994.

[23] M. Bichsel, "Segmenting Simply Connected Moving Objects in a Static Scene," Trans. Pattern Analysis and Machine Intelligence, vol. 16, no. 11, pp. 1,138–1,142, Nov. 1994.

[24] C. R. Wren, A. Azarbayejani, T. Darrell, and A. P. Pentland, "Pfinder: Real-Time Tracking of the Human Body", Trans. Pattern Analysis and Machine Intelligence, vol. 19, 1997.

[25] C. Stauffer, W. Grimso, "Adaptive background mixture models for real-time tracking" ,CVPR, 1998.

[26] L. Li, W. Huang, I. Y.H. Gu, Q. Tian, "Foreground Object Detection from Videos Containing Complex Background", ACM, 2003.

[27] D. da Silva Pires, R. Cesar-Jr.,"Tracking and Matching Connected Components from 3D Video", CVPR, 2005.

[28] C. Shan, Y. Wei, T. Tan, F. Ojardias, "Real Time Hand Tracking by Combining Particle Filtering and Mean Shift", Proceedings of the Sixth IEEE International Conference on Automatic Face and Gesture Recognition, 2004.